%% file: main.tex
\documentclass[sigconf, screen]{acmart}
\AtBeginDocument{%
  }

\setcopyright{acmlicensed}
\copyrightyear{2018}
\acmYear{2018}
\acmDOI{XXXXXXX.XXXXXXX}
\acmConference[Conference acronym 'XX]{Make sure to enter the correct
  conference title from your rights confirmation email}{June 03--05,
  2018}{Woodstock, NY}
\acmISBN{978-1-4503-XXXX-X/2018/06}

\usepackage{multirow}
\usepackage{xcolor}
\usepackage{color}
\usepackage{colortbl}
\usepackage{booktabs}
\usepackage{subfigure}
\usepackage{graphicx}
\usepackage[most]{tcolorbox}
\usepackage{hyperref}
\usepackage{adjustbox}
\usepackage{url}  
\usepackage{subcaption}
\newcommand{\benchmark}{\textsc{M3STR}}

\definecolor{green0}{RGB}{190, 245, 163}
\newcommand{\SR}{\cellcolor{green0}}
\definecolor{green1}{RGB}{118, 189, 83}
\newcommand{\UR}{\cellcolor{green1}}
\definecolor{green2}{RGB}{119,153,119}
\newcommand{\SSR}{\cellcolor{green2}}
\definecolor{green3}{RGB}{0,100,0}

\begin{document}

\title{Abstractive Visual Understanding of Multi-modal Structured Knowledge: A New Perspective for MLLM Evaluation}

\author{Yichi Zhang}
\email{zhangyichi.each@zju.edu.cn}
\affiliation{%
  \institution{Zhejiang University}
  \city{Hangzhou}
  \state{Zhejiang}
  \country{China}
}

\author{Zhuo Chen}
\email{zhuo.chen@zju.edu.cn}
\affiliation{%
  \institution{Zhejiang University}
  \city{Hangzhou}
  \state{Zhejiang}
  \country{China}
}

\author{Lingbing Guo}
\email{lbguo@tju.edu.cn}
\affiliation{%
  \institution{Tianjin University}
  \city{Tianjin}
  \state{Tianjin}
  \country{China}
}

\author{Yajing Xu}
\email{xyjing@zju.edu.cn}
\affiliation{%
  \institution{Zhejiang University}
  \city{Hangzhou}
  \state{Zhejiang}
  \country{China}
}

\author{Min Zhang}
\email{minzhang@suda.edu.cn}
\affiliation{%
  \institution{Harbin Institute of Technology}
  \city{Shenzhen}
  \state{Guangdong}
  \country{China}
}

\author{Wen Zhang}
\email{zhang.wen@zju.edu.cn}
\affiliation{%
  \institution{Zhejiang University}
  \city{Hangzhou}
  \state{Zhejiang}
  \country{China}
}

\author{Huajun Chen}
\email{huajunsir@zju.edu.cn}
\affiliation{%
  \institution{Zhejiang University}
  \city{Hangzhou}
  \state{Zhejiang}
  \country{China}
}

\renewcommand{\shortauthors}{Trovato et al.}
\renewcommand{\shorttitle}{Abstractive Visual Understanding of Multi-modal Structured Knowledge}

\begin{abstract}
    Multi-modal large language models (MLLMs) incorporate heterogeneous modalities into LLMs, enabling a comprehensive understanding of diverse scenarios and objects. Despite the proliferation of evaluation benchmarks and leaderboards for MLLMs, they predominantly overlook the critical capacity of MLLMs to comprehend world knowledge with structured abstractions that appear in visual form. To address this gap, we propose a novel evaluation paradigm and devise {\benchmark}, an innovative benchmark grounded in the \underline{\textbf{M}}ulti-\underline{\textbf{M}}odal \underline{\textbf{M}}ap for \underline{\textbf{STR}}uctured understanding. This benchmark leverages multi-modal knowledge graphs to synthesize images encapsulating subgraph architectures enriched with multi-modal entities. {\benchmark} necessitates that MLLMs not only recognize the multi-modal entities within the visual inputs but also decipher intricate relational topologies among them. We delineate the benchmark's statistical profiles and automated construction pipeline, accompanied by an extensive empirical analysis of 26 state-of-the-art MLLMs. Our findings reveal persistent deficiencies in processing abstractive visual information with structured knowledge, thereby charting a pivotal trajectory for advancing MLLMs' holistic reasoning capacities. Our code and data are released at \url{https://github.com/zjukg/M3STR}.
\end{abstract}



\maketitle

\input{chapters/1-introduction}
\input{chapters/2-relatedworks}
\input{chapters/3-preliminary}
\input{chapters/4-method}
\input{chapters/5-experiments}
\input{chapters/6-conclusion}

\bibliographystyle{ACM-Reference-Format}
\bibliography{sample-base}

\end{document}

%% file: chapters/1-introduction.tex
\begin{figure}[]
  \centering
  \includegraphics[width=\linewidth]{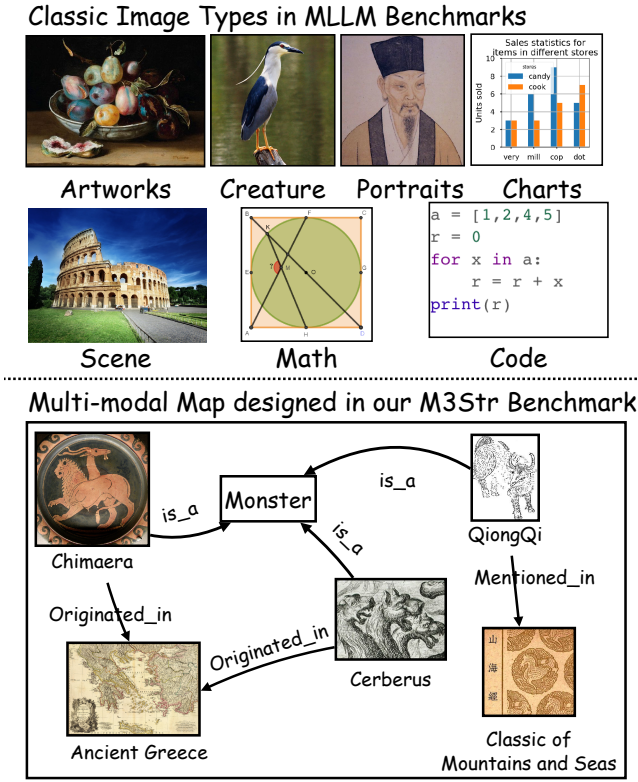}
  \vspace{-16pt}
  \caption{A simple case to demonstrate the difference of image data between {\benchmark} and existing benchmarks. We visualize KGs with higher-level abstraction as the image inputs. Therefore, our data consists of more complex multi-modal information about entities and their relational typology.}
  \label{figure::introduction}
  \vspace{-16pt}
\end{figure}
\section{Introduction}
Multi-modal large language model (MLLM) \cite{MLLM-survey, MLLM-connector} represent an evolutionary leap beyond traditional large language models (LLMs) \cite{llama3} with multi-modal content understanding and generation capabilities, which is now the forefront in the multimedia community. By bridging the gap of heterogeneous multi-modal representation spaces with connectors \cite{MLLM-connector}, MLLMs like Qwen2.5-VL \cite{qwen2.5-VL}, LLaVA \cite{llava-1.5} can process different modality information and solve diverse multi-modal tasks.

\par As illustrated in Figure \ref{figure::introduction}, numerous systematically designed benchmarks \cite{MMBench, MMT-Bench, MME-survey} have emerged to evaluate the multi-dimensional capabilities of the MLLMs, which encompass images about natural scenarios, portraits, various creatures, and objects captured from the real world. Some other MLLM benchmarks also build up on synthetic images with mathematics \cite{Math-VISTA}, code, and charts \cite{ChartQA1}. Nevertheless, existing benchmarking paradigms conspicuously overlook a pivotal dimension of MLLM proficiency: the comprehension and interpretation of visual content containing highly abstract structured knowledge. Such structured representations—exemplified by mind maps and knowledge graphs (KGs), pervade daily life. Distinctive from conventional imagery, these constructs simultaneously encode concrete visual entities and intricate relational semantics. MLLMs must not only recognize the entities but also decipher the relational topology structures and abstractive connections between them, which would be abstractive and difficult to understand. This competency proves imperative for MLLMs as it fundamentally reflects their capacity to internalize structured world knowledge through visual modalities, which is a foundational requirement for developing artificial general intelligence that mirrors human conceptual understanding capability.

\par To address the gap in current research, we propose a new perspective for MLLM evaluation called abstractive visual understanding of structured knowledge. We build a novel benchmark {\benchmark}, designated as \underline{\textbf{M}}ulti-\underline{\textbf{M}}odal \underline{\textbf{M}}ap for \underline{\textbf{STR}}uctured undetstanding. {\benchmark} leverages multi-modal knowledge graphs (MMKG) as the data source, which encapsulates diverse structured human knowledge in the image inputs. Figure \ref{figure::introduction} presents a simple demonstration of the image samples, referred to as a multi-modal map in this paper. We design three types of tasks in {\benchmark} benchmark called count, detection, and completion. These tasks evaluate different aspects of MLLMs' understanding of multi-modal maps. The three tasks are further extended into several granular subtasks that specifically target entity and relation understanding. We propose a pipeline to synthesize the multi-modal maps, integrating both structured knowledge and multi-modal content. First, we sample subgraph instances from a large-scale MMKG. Then, we make task-specific modifications to these subgraphs. Finally, we transform the subgraph with multi-modal information into an image using a visualization API. Additionally, we design a task-specific prompt template to guide the MLLMs. \textbf{The core difference between our data and existing benchmarks is that the images contain more abstractive entity information and their relational topology.} We conduct a comprehensive evaluation of 26 mainstream MLLMs. In addition to an evaluation leaderboard, we perform a series of exploratory experiments to examine whether the entity-specific multimodal information present in the multi-modal map impacts final predictions. Based on our experiments, we draw several key insights, revealing that \textbf{current MLLMs still struggle to comprehend high-level abstractions in the visual modality}. These findings highlight significant capacity deficits in MLLMs and underscore the potential for future improvements, suggesting a new direction for community development. In summary, our contribution to this paper is threefold:
\par (1). We propose a new perspective to evaluate MLLMs' abstractive visual understanding capability on structured knowledge, which has not been explored before.
\par (2) We construct a new benchmark called {\benchmark} with a new pipeline, consisting of diverse task types and data distribution.
\par (3). We conduct a comprehensive evaluation on {\benchmark} with 26 mainstream MLLMs. Our interesting findings indicate that current MLLMs are struggling with abstractive content understanding.
\input{chapters/4-dataset}

%% file: chapters/4-dataset.tex
\begin{figure*}[]
  \centering
  \includegraphics[width=0.9\linewidth]{pictures/dataset_overview.pdf}
  \vspace{-24pt}
  \caption{An overview of the constructed {\benchmark} benchmark. We show the compositional structure of {\benchmark} with its three task types and the corresponding subtasks in the (a) part. We also list the basic formulation and expected outputs of the tasks in the (b) part. The distrbution of the entity number in each task is presented in the (c) part.}
  \label{figure::dataset}
  \vspace{-8pt}
\end{figure*}

%% file: chapters/2-relatedworks.tex
\section{Related Works}
\subsection{Multi-modal Large Language Models}
Multi-modal large language models (MLLMs) \cite{MLLM-survey} are a type of large language models (LLMs) \cite{llama3} with multi-modal content comprehension and reasoning capabilities. Current LLMs can support diverse types of modalities such as images \cite{qwen2.5-VL}, audio \cite{AudioGPT}, videos \cite{Videollama}, etc. As a general paradigm, the multi-modal contents are encoded with modality encoders (e.g., ViT \cite{ViT}, CLIP \cite{CLIP}) and project into the textual representation space of LLMs with a specific connector \cite{MLLM-connector}. The connector can be a simple linear layer \cite{llava} or complex neural networks like Q-former \cite{minigpt4} or MoE \cite{onellm}, aiming to bridge the modality gap. Mainstream MLLMs usually focus on their visual understanding and reasoning capabilities. For example, LLaVA series \cite{llava}, MiniCPM-V series \cite{minicpmv}, and Qwen-VL series \cite{qwen2.5-VL} demonstrates outperforming results on diverse tasks.

\subsection{MLLM Capability Evaluation}
As a wide variety of MLLMs have been proposed, how to evaluate various capabilities of MLLMs \cite{MME-survey} has become an important research topic. Diverse benchmark datasets are constructed to evaluate the capabilities of MLLMs from different perspective: general multi-modal understanding \cite{MMT-Bench, MMBench}, multi-disciplinary \cite{ScienceQA, MMLU}, multi-lingual \cite{CMMU}, fine-grained perception \cite{MMVP-Finegrained}, chart understanding \cite{ChartQA1, ChartQA2}, mathematical reasoning \cite{Math-VISTA}, domain-specific application \cite{MedVQA} etc. These benchmarks are constructed based on different motivations and pipelines and are used to assess the capabilities of MLLMs with diverse task formulations. The image-text data in these benchmarks are mostly collected from existing datasets or the Internet with manual or automatic annotation \cite{MME-survey}. Most image data is captured from the real world, such as natural landscapes, buildings, portraits, and various objects, and some images contain more complex and abstract human-generated information, such as charts and mathematical questions. We argue that the structured knowledge in the visual modality is active on the Internet (e.g., K12 education, enterprise workflows, personal notes), which consists of structured abstract relational topology and integration of multi-dimensional information. Understanding these abstractive images is also a necessary capability of modern MLLMs, and we are the first to explore it in this paper.


\subsection{Multi-modal Knowledge Graphs}
Knowledge graph (KG) \cite{KGSURVEY-TPAMI24} is a kind of semantic network that models factual knowledge with structured triples as \textit{(head entity, relation, tail entity)}. The massive triples can form a large semantic graph in which entities are the nodes and relations are the edges.
Multi-modal knowledge graph (MMKGs) \cite{MMKG-survey} is a special type of KG that incorporates multi-modal contents about the entities and relations to make semantic-rich structured knowledge representation. Traditional MMKG \cite{MMKG} usually consists of textual descriptions and visual images of entities while recent new MMKGs \cite{TIVA,KuaiPedia} incorporate more diverse modalities such as audio and video. Nowadays, integrating KG and LLM \cite{KG+LLM} becomes an important direction for the research community. Many works \cite{KoPA-MM24,GNP-AAAI24,SUBARU,LLMKG1} attempt to incorporate the factual knowledge in KGs and MMKGs into LLMs and MLLMs to enhance the correctness and trustworthiness of the generated answer. In this paper, we employ MMKGs as the knowledge source to build benchmarks for abstractive visual understanding and reasoning evaluation for MLLMs.

%% file: chapters/3-preliminary.tex

\section{Methodology}
\label{section::method}
\subsection{Preliminary}
\label{sec:preliminary}
\subsubsection{MLLM}
A general MLLM that supports image input can be denoted as $\mathcal{M}$, which can generate an answer $\mathcal{A}^*$ based on the image-text pair input that maximizes the next-token prediction probability. This process is denoted as:
\begin{equation}
    \mathcal{A}^*=\arg\max_{\mathcal{A}} P_{\mathcal{M}} (\mathcal{A}\mid \mathcal{I}, \mathcal{Q})
\end{equation}
Where $\mathcal{I}, \mathcal{Q}$ represent the input image and text prompt (a question) related to the image. Our work aims to build a benchmark dataset consisting of image-text pairs to evaluate the abstractive structured knowledge understanding and reasoning capability of the MLLMs.
\subsubsection{MMKG}
To build the benchmark, we use MMKG as our original data source. An MMKG can be denoted as $\mathcal{KG}=(\mathcal{E}, \mathcal{R}, \mathcal{T}, \mathcal{V}, \mathcal{D})$ where $\mathcal{E},\mathcal{R}$ represent the entity set and the relation set. $\mathcal{T}=\{(h, r, t)\mid h, t\in\mathcal{E}, r\in\mathcal{R}\}$ is the triple set and $\mathcal{V}, \mathcal{D}$ are the visual image / textual description sets for the entities and relation.

%% file: chapters/4-method.tex
\subsection{The General Motivation}
\input{chapters/4-pipeline}
In this paper, we aim to design a new benchmark to measure current MLLMs from a new perspective: the ability to comprehend structured knowledge that contains a high degree of abstraction. MMKGs are used as a data source because MMKGs not only contain a large amount of knowledge but also present a semi-structured organizational form. To visually interpret the information in a KG, it is essential for MLLMs not only to understand each specific entity but also to grasp the relational topology between them. Unlike natural scenes where objects typically follow physical and spatial laws, the visual representations of KGs form a more abstract and complex graph structure with multi-modal entity information and relational edges.
Next, we describe how to construct such a benchmark and then report the experimental results over the benchmark.

\subsection{Benchmark Overview}
We present the compositional structure of our \underline{\textbf{M}}ulti-\underline{\textbf{M}}odal \underline{\textbf{M}}ap for \underline{\textbf{STR}}uctured understanding of MLLMs ({\benchmark} for short, pronounced as 'monster') in Figure \ref{figure::dataset}(a). We designed three types of tasks to varying degrees of abstractive visual understanding ability from evaluating MLLMs, which are called \textbf{count, detection, and completion.} {\benchmark} provides an image-text pair as the input, which is the multi-modal map and text prompt respectively. Furthermore, each type consists of several (2-3) subtasks, each with a specific focus, which we summarize below:
\par (1). \textbf{Task1: Count}: This task requires MLLMs to count the number of entities and relations in it. This task falls under \textbf{coarse-grained object recognition} and evaluates MLLM's superficial recognition of MMKGs.
\par (2). \textbf{Task2: Detection}: This task requires MLLMs to detect whether there is a factual anomaly in the given subgraph of MMKG and answer "Yes" or "No". To answer correctly, the MLLM must demonstrate a higher level of judgment about the commonsense information embedded in the MMKG. By understanding the content of the MMKG, the model must decide whether the subgraph is reasonable or contains errors, using a binary classification output.
\par (3). \textbf{Task3: Completion}: This task requires the model to predict a missing entity/relation in the given MMKG with local contexts. The neighborhood information would be provided with the answer being masked. Such a task setting is similar to knowledge graph completion (KGC) \cite{MyGO} a classic research topic in the KG community. Successfully predicting the missing component demonstrates the MLLM's ability to understand the current MMKG and make simple inferences.

\par We present the basic formulation of each task in Figure \ref{figure::dataset}, showing their task types and expected output formats. Some concrete examples of the three types of data are presented in Figure \ref{figure::model}. Note that all the mentioned three different task types require a subgraph captured from MMKG as the input. In our setting, the subgraph would be translated into an image $\mathcal{I}$  with visualization APIs as the input, while a task-specific question prompt $Q$ to inform and guide the MLLMs to read the image input and give an answer $\mathcal{A}$. This setup allows us to evaluate the MLLMs' ability to understand and reason about the given multi-modal subgraph.

\par For each task type, we also introduce several subtasks for more granular evaluation. Since entities and relations are the fundamental components of MMKGs, we define 2 subtasks for each task that focus on either entities or relations. This means that, in each subtask, the entities or relations in the input $\mathcal{I}$ are the primary focus. For Task 1, the subtask types are \textbf{entity count} and \textbf{relation count}, where the goal is to count the number of entities or relations in $\mathcal{I}$, respectively. For Task 2, the subtasks are \textbf{entity detection} and \textbf{relation detection}, where the model must determine whether there are any suspicious entities or relations in the MMKG. For Task 3, the missing object to be predicted will either be an entity or a relation. The specific task setting is reflected in the corresponding question template, $\mathcal{Q}$. Besides, we have an extra setting called mix for the detection task, which would inform the MLLMs to discriminate arbitrary anomalies with no scope constraints. This results in a more generalized detection task. In this way, we constructed 3 different tasks with a total of 7 different subtasks.

\subsection{Construction Process of {\benchmark}}
The detailed construction process of the {\benchmark} benchmark consists of three distinct steps: (1) subgraph sampling, (2) data instance construction, and (3) visual translation. We will introduce the details of the three steps in the following sections. The original MMKG data used in our {\benchmark} benchmark is FB15K-237 \cite{MMKG}, which is a classic encyclopedic KG extracted from Freebase \cite{Freebase-SIGMOD08} with image extensions. The overall pipeline for constructing the {\benchmark} is presented in Figure \ref{figure::model}.
\subsubsection{Subgraph Sampling}
To evaluate the abstractive visual understanding capability of MLLMs, we need data instances in the form of $(\mathcal{I}, \mathcal{Q}, \mathcal{A})$, which are the input image, text prompt, and the golden answer respectively. Moreover, $\mathcal{I}$ represents a small MMKG with contextual information. Therefore, we extract $\mathcal{G}'$, subgraph from the given MMKG $\mathcal{KG}$, which can be denoted as:
\begin{equation}
    \mathcal{G}'=(\mathcal{E}', \mathcal{R}', \mathcal{T}'), \quad \mathcal{E}'\subsetneqq \mathcal{E}, \mathcal{R}'\subsetneqq \mathcal{R}
\end{equation}
\begin{equation}
    \mathcal{T}'=\{ (h,r,t)\mid h,r \in\mathcal{E}', r\in \mathcal{R}', (h,r,t)\in \mathcal{T}\}\subsetneqq\mathcal{T}
\end{equation}
where $\mathcal{E}', \mathcal{R}'$ are the entity and relation subsets of the subgraph $\mathcal{G}'$. $\mathcal{T}'$ represents the sampled triples in the subgraph $\mathcal{G}'$. All of the sets are the corresponding subsets of the original KG. We perform random sampling on the given MMKG with a random sample. Given a start entity $e$ and targeted size $K$, the subgraph is obtained by:
\begin{equation}
    \mathcal{G}'\leftarrow \texttt{RandomSampler}(e, K, \mathcal{KG})
\end{equation}
where $\texttt{RandomSampler}(\cdot)$ would start from the entity $e$ with iterative depth-first or breadth-first search. The process continues until a subgraph with $K$ entities is obtained while preserving all relations associated with those entities. The subgraphs will be further modified in the next steps.
\subsubsection{Data Instance Construction}
Note that different tasks have their specific formulation and setting. Therefore, we make further modifications on $\mathcal{G}'$ to transform it into a task-specific data prototype. This process can be denoted as:
\begin{equation}
    \mathcal{G}'', \mathcal{A}\leftarrow \texttt{Modifier}_{task}(\mathcal{G}')
\end{equation}
where $\texttt{Modifier}_{task}(\cdot)$ is a task-specific modification operation and $\mathcal{A}$ is the golden answer for the current data. 
\par (1). \textbf{For the count task}, $\texttt{Modifier}(\cdot)$ does not modify the sampled graph. Instead, it simply counts the number of entities or relations as the answer. This is because object counting can be performed directly on a raw subgraph without further modifications.
\par (2). \textbf{For the detection task}, $\texttt{Modifier}(\cdot)$ modifies an entity or relation in the graph by replacing it with another entity that does not exist in the current subgraph. This introduces noise and factual errors into the subgraph. The modification is performed with a certain probability. For these modified subgraphs, the golden answer would be "Yes", indicating the presence of an anomaly, while for unchanged subgraphs, it would be "No". This process is similar to the negative sampling \cite{DBLP:conf/nips/TransE} in contrastive learning. We construct negative instances with manual disruption. The overall ratio of positive to negative samples is controlled to be 1:1.
\par (3). \textbf{For the completion task}, $\texttt{Modifier}(\cdot)$ randomly masks one entity or relation in the given subgraph with a dummy placeholder. The masked object is then treated as the golden answer, while four other potential options are sampled from the MMKG. This allows the model to infer the missing object based on the context.
\subsubsection{Visual Translation}
Finally, we translate the modified subgraph $\mathcal{G}''$ into visual modality as an image, which is implemented by visualization APIs. This process can be denoted as:
\begin{equation}
    \mathcal{I}\leftarrow \texttt{Visualize}(\mathcal{G}'', \mathcal{V}, \mathcal{D})
\end{equation}
In practice, we leverage GraphViz \footnote{\url{https://graphviz.org/}}, a famous open-source library for graph visualization. Note that both the images of the entities in subgraph $\mathcal{G}''$ and the textual descriptions of the entities and relations are utilized as inputs when generating the visual representation. Therefore, we obtain a semantic-rich image $\mathcal{I}$ consisting of the three modality information of subgraph $\mathcal{G}''$: \textbf{the graph structure, the visual information of entities, and the text description of entities}. These modalities coexist in the final visual form within a single image. Besides, we prepare a task-specific prompt template for each subtask, which will serve as the question $\mathcal{Q}$. For the completion task, the input prompt consists of extra information on the five options to guide the MLLMs to make a choice. In this way, we construct the data instances $(\mathcal{I}, \mathcal{Q}, \mathcal{A})$ for each task in our {\benchmark} benchmark, aiming to evaluate the abstractive visual understanding and reasoning capability of MLLMs. Some demonstrations of the generated image data and question prompts are presented in Figure \ref{figure::model}(b)(c) for an intuitive view.
Next, we will show the detailed evaluation protocol and results.

\subsection{Evaluation Protocol}
As we mentioned in the previous section, the unified format of each data instance would be $(\mathcal{I}, \mathcal{Q}, \mathcal{A})$. During the evaluation process, we would inform the MLLMs with image $\mathcal{I}$ and the question prompt $\mathcal{Q}$ to obtain the output $\mathcal{A}^*$. For a certain subtask $t$, the final score $\mathcal{S}_t$ of MLLMs would be the accuracy of generated results and the golden answer. We further define an \textbf{overall score} $\mathcal{S}$ of the MLLMs which is the \textbf{average of 7 subtasks} to present the general capability of MLLMs on the abstractive visual reasoning task.

%% file: chapters/4-pipeline.tex
\begin{figure*}[]
  \centering
  \includegraphics[width=\linewidth]{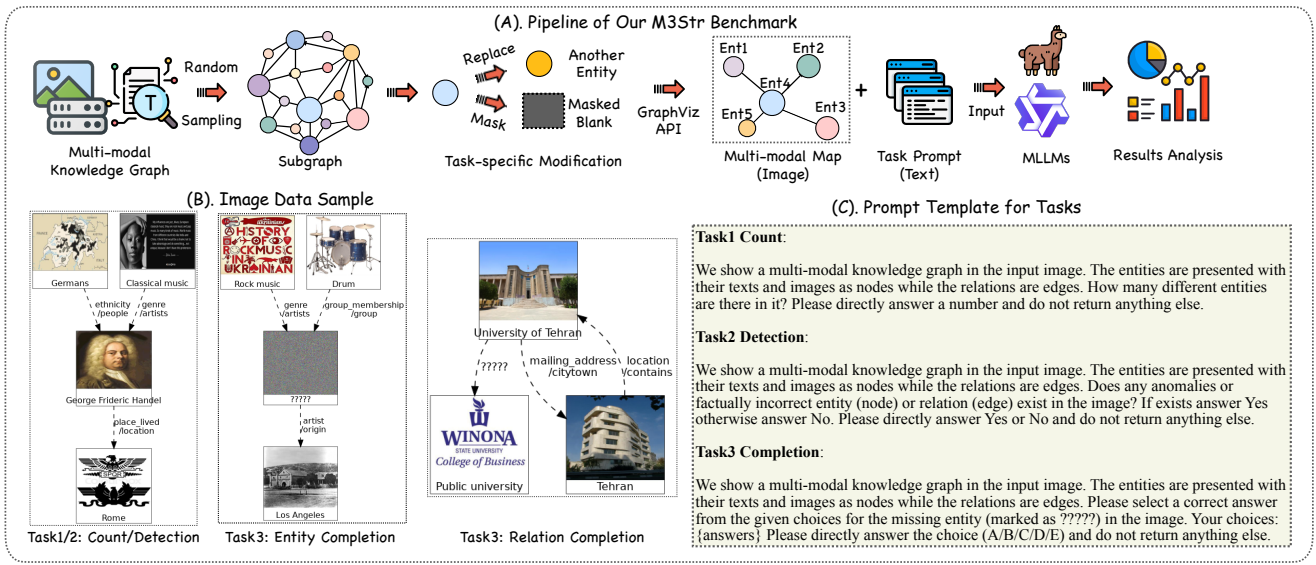}
  \vspace{-12pt}
  \caption{The construction pipeline, image data sample, and prompt template of {\benchmark} benchmark. For detection task, the image data sample may be negative sample with noise. For completion task, the targeted entity / relation is masked as ``?????".}
  \label{figure::model}
  \vspace{-8pt}
\end{figure*}

%% file: chapters/5-experiments.tex
\section{Experiments and Evaluation}
\label{sec:experiments}
In this section, we will present the detailed experiment and evaluation results of the proposed {\benchmark} on diverse MLLMs. Our analysis addresses the following three \textbf{research questions (RQ)}:
\par \textbf{RQ1}: How do mainstream MLLMs perform on {\benchmark}'s three tasks? Which model reached state-of-the-art on the task of abstract visual reasoning?
\par \textbf{RQ2}: Can MLLMs fully understand the multi-modal maps in the visual context? Which modality contributes to the final output?
\par \textbf{RQ3}: How are the answers output by MLLMs distributed, and are there certain biases and flaws in them?
\input{chapters/5.2-table}
\subsection{Experiment Settings}

\subsubsection{Employed MLLMs}
In our experiments, we evaluate 23 mainstream open-source MLLMs with the constructed {\benchmark} benchmark, including InstructBLIP \cite{InstructBLIP}, LLaVA \cite{llava, llava-1.5}, Chameleon \cite{Chameleon}, MiniCPM-V \cite{minicpmv}, Deepseek-VL \cite{deepseekvl}, Deepseek-VL2 \cite{deepseekvl2}, InternVL-2.5 \cite{InternVL2.5}, Phi-vision \cite{phi-3v}, Qwen2-VL \cite{qwen2vl}, Qwen2.5-VL \cite{qwen2.5-VL}. We experimented on all these series of MLLMs of different sizes from 1B to 72B. We also report the performance on 3 famous API-based LLMs that support multi-modal input: GPT-4V-turbo \cite{gpt4}, GPT-4o-mini \cite{openai2024gpt4ocard}, and Gemini-2.5-pro \footnote{\url{https://gemini.google.com/}} for comprehensive evaluation.

\subsubsection{Evaluation Details}
We first download the pre-trained MLLMs weight from HuggingFace Model Hub \footnote{\url{https://huggingface.co/models}} and ModelScope \footnote{\url{https://www.modelscope.cn/}}. We implement the inference code of MLLMs with two famous open-source projects called transformers \cite{transformers} and vLLM \cite{vLLM}. We thank the maintainers of these open-source projects and platforms, which have been very helpful to our research. In our evaluation, we set the temperature to 0.0 for the reproducibility of our results. For the open-source MLLMs supported by vLLMs, we employ guided decoding to keep results in the candidate range. All the experiments are conducted on a Linux server with NVIDIA A800 GPUs.

\input{chapters/5.2-mainexp}

\input{chapters/5.3-ablation}

\input{chapters/5.4-case}

%% file: chapters/5.2-table.tex
\begin{table*}[]
\caption{Main evaluation results on MLLMs with different architectures and sizes. We present the task-wise accuracy results (\%) and the overal ranking of 23 open-source MLLMs and 3 API-based MLLMs.}
\label{table::main}
\resizebox{0.9\textwidth}{!}{
\begin{tabular}{cl|cc|ccc|cc|cc}
\toprule
\multicolumn{2}{c|}{\multirow{2}{*}{\textbf{Model Information}}} & \multicolumn{2}{c|}{\textbf{Task 1: Count}} & \multicolumn{3}{c|}{\textbf{Task 2: Detection}} & \multicolumn{2}{c|}{\textbf{Task 3: Completion}} & \multirow{2}{*}{\textbf{Overall}} & \multirow{2}{*}{\textbf{Rank}} \\
\multicolumn{2}{c|}{} & \textbf{Entity} & \textbf{Relation} & \textbf{Entity} & \textbf{Relation} & \textbf{Mix} & \textbf{Entity} & \textbf{Relation} &  &  \\
\midrule
\multicolumn{2}{c|}{\textbf{Random Choice}} & 11.11 & 11.11 & 66.66 & 83.66 & 50.00 & 20.00 & 20.00 & 37.51 & 22 \\
\midrule
\multicolumn{11}{c}{\textit{\textbf{Open-source MLLMs}}}\\
\midrule
\multirow{2}{*}{\textbf{InstructBLIP}} & Vicuna-7B & 12.56 & 11.15 & 33.35 & 16.49 & 49.84 & 43.77 & 33.07 & 28.60 & 25 \\
 & Vicuna-13B & 12.61 & 11.22 & 33.35 & 16.49 & 49.84 & 37.35 & 28.67 & 27.08 & 26 \\
\midrule
\multirow{3}{*}{\textbf{LLaVA}} & 1.5-7B & 14.49 & 11.26 & 66.64 & \SR 83.51 & 50.15 & 57.46 & 66.03 & 49.93 & 18 \\
 & 1.6-Vicuna-7B & 16.21 & 16.04 & 66.64 & 83.48 & 50.18 & 64.12 & 72.13 & 52.69 & 15 \\
 & 1.1-Llama3-8B & 21.62 & 17.89 & 33.86 & 16.74 & 50.01 & 52.98 & 59.73 & 36.12 & 23 \\
\midrule
\textbf{Chameleon} & 7B & 12.12 & 10.75 & 33.49 & 16.54 & 50.18 & 18.97 & 20.38 & 23.20 & 27 \\
\midrule
\multirow{3}{*}{\textbf{MiniCPM-V}} & V2.0-2.8B & 28.53 & 17.22 & 66.64 & \SR 83.51 & 50.15 & 66.47 & 79.05 & 55.94 & 9 \\
 & V2.5-8B & 35.46 & 19.28 & 66.79 & 81.98 & 50.27 & 65.20 & 80.42 & 57.06 & 7 \\
 & V2.6-8B & \SR 50.56 & 24.16 & \SR 67.36 & 77.73 & 51.47 & 84.51 & 84.64 & 62.92 & 4 \\
\midrule
\multirow{4}{*}{\textbf{DeepSeek}} & VL-1.3B & 21.72 & 27.60 & 66.56 & 82.99 & 50.10 & 53.55 & 64.01 & 52.36 & 16 \\
 & VL-7B & 29.02 & 24.52 & 66.59 & 83.42 & 50.10 & 65.81 & 73.55 & 56.14 & 8 \\
 & VL2-3B & 26.04 & 28.42 & 66.64 & 83.50 & 50.15 & 53.61 & 71.18 & 54.22 & 12 \\
 & VL2-16B & 32.13 & 35.43 & 41.25 & 36.80 & 50.44 & 40.92 & 69.90 & 43.84 & 21 \\
\midrule
\multirow{2}{*}{\textbf{Intern-VL}} & 2.5-1B & 22.61 & 16.73 & 66.39 & 67.73 & 50.21 & 54.02 & 75.08 & 50.40 & 17 \\
 & 2.5-8B & 42.61 & 28.15 & 58.69 & 25.94 & 53.49 & 80.43 & 84.08 & 53.34 & 13 \\
\midrule
\multirow{2}{*}{\textbf{Phi}} & 3-vision-4.2B & 24.61 & 21.44 & 66.59 & 83.45 & 50.19 & 65.38 & 78.89 & 55.79 & 11 \\
 & 3.5-vision-4.2B & 36.31 & 21.41 & 66.64 & 83.51 & 50.15 & 60.01 & 73.22 & 55.89 & 10 \\
\midrule
\multirow{3}{*}{\textbf{Qwen2-VL}} & 2B-Instruct & 16.86 & 16.39 & 66.64 & \SR 83.51 & 50.15 & 59.89 & 76.66 & 52.87 & 14 \\
 & 7B-Instruct & 26.84 & 20.68 & 66.73 & 81.49 & 50.99 & \SR 84.59 & 85.69 & 59.57 & 6 \\
 & 72B-Instruct & \UR 65.76 & 27.04 & \SSR 71.18 & 72.27 & \UR 60.98 & \SSR 93.17 & \UR 88.87 & \UR 68.47 & \UR 2 \\
\midrule
\multirow{3}{*}{\textbf{Qwen2.5-VL}} & 3B-Instruct & 38.75 & 22.71 & 66.64 & \SSR 83.51 & 50.15 & 79.50 & 84.29 & 60.79 & 5 \\
 & 7B-Instruct & 44.81 & \SSR 32.88 & 66.64 & \UR 83.51 & 50.16 & 82.42 & \SR 86.06 & \SR 63.78 & \SR 3 \\
 & 72B-Instruct & \SSR 70.94 & 27.82 & \UR 70.38 & 64.75 & \SSR 65.61 & \UR 91.18 & \SSR 89.46 & \SSR 68.59 & \SSR 1 \\
\midrule
\multicolumn{11}{c}{\textit{\textbf{API-based MLLMs}}}\\
\midrule
\multirow{2}{*}{\textbf{OpenAI}} & GPT-4V-Turbo & 42.50 & \SR 28.02 & 41.04 & 22.35 & 52.24 & 77.19 & 74.37 & 48.24 & 20 \\
 & GPT-4o-mini & 40.55 & \UR 31.34 & 57.16 & 46.35 & \SR 54.95 & 53.79 & 61.38 & 49.36 & 19 \\
\midrule
{\textbf{Google}} & Geimini-2.5-pro & 23.55 & 8.71 & 40.96 & 37.75 & 53.72 & 21.05 & 23.15 & 29.84 & 24 \\
\bottomrule
\end{tabular}
}
\vspace{-8pt}
\end{table*}

%% file: chapters/5.2-mainexp.tex
\input{chapters/5.3-figure}
\subsection{Main Experiment Results (RQ1)}
The main evaluation results of 26 mainstream MLLMs are summarized in Table \ref{table::main}. We report the task-wise performance alongside the overall score and final ranks for the MLLMs. Analysis of the results reveals several noteworthy findings from multiple perspectives.

\paragraph{ \textbf{Perspective 1: Open-source MLLMs achieve state-of-the-art performance on {\benchmark}.}}  As indicated in Table \ref{table::main}, Qwen2.5-VL-72B achieves state-of-the-art overall results. The Qwen models are scored brightly in the results, while other open-source MLLMs like InternVL, Phi, and DeepSeek-VL also exhibit competitive performance. However, famous closed-source APIs like GPT-4V / GPT-4o and Gemini only get a relatively low ranking. Notably, Qwen models occupy four of the top five positions, demonstrating clear dominance on this benchmark. For instance, GPT-4V-Turbo ranks 20th among the 26 evaluated MLLMs, underperforming relative to most open-source counterparts. This trend indicates that open-source MLLMs generally excel over closed-source models in comprehending and reasoning over visual MMKGs.

\paragraph{ \textbf{Perspective 2: Scaling law still works in abstractive visual understanding and reasoning.}} Scalind law \cite{scaling_law} reveals that larger models tend to have better performance after sufficient training, which is validated in our experiments, although {\benchmark} holds generally new tasks. In the Qwen2-VL and Qwen2.5-VL model series, as model size increases (2B/3B->7B->72B), the task-wise performance exhibits a significant improvement. For instance, in Task 1 (Entity Count), Qwen2.5-7B yields a 1.16× improvement over Qwen2.5-3B, while Qwen2.5-72B achieves a 1.58× gain compared to Qwen2.5-7B. These results highlight the gain of model scaling, with a particularly notable improvement observed between the 7B and 72B parameter scales. Furthermore, generational advancements in MLLMs consistently drive significant performance improvements, reflecting effective optimizations in model architecture, data pipelines, training strategies, and overall methodology. This trend is evident in the stable and marked progress across successive iterations of models such as MiniCPM, the transition from Phi-3 to Phi-3.5, and the evolution from Qwen2-VL to Qwen2.5-VL.

\paragraph{ \textbf{Perspective 3: Current MLLMs still struggle to abstract visual understanding.}} From the results of the experiments, it is clear that current MLLMs, especially smaller MLLMs, reveal substantial limitations in abstract visual understanding. We can find that the performance of many models on some specific subtasks approximates purely random guessing. For instance, Task 1 (Entity Count), the simplest task requiring only counting of distinct entities and relations within the input image, yields accuracies below 30\% for half of the evaluated MLLMs. This indicates a fundamental deficit in basic visual perception; current models frequently fail to accurately quantify even the core elements and the relational topology of the structured knowledge described by the images.
\par Furthermore, Task 2 (Anomaly Detection), which demands identifying inconsistencies within the MMKG, presents a more severe challenge. Surprisingly, nearly all MLLMs perform at near-random levels or even worse. In the mixed experimental setting, only Qwen2/2.5-VL-72B and API-based models demonstrate better performance than random choice, yet their accuracy remains suboptimal. This reveals that MLLMs fundamentally fail to detect entity/relation-level errors and anomalies within input MMKGs, thereby rendering them incapable of making accurate judgments. Paradoxically, in contrast to the pervasive hallucination issues in MLLMs, these models demonstrate relatively stronger performance on Task 3 (presented in multiple-choice format), with many achieving over 80\% accuracy under zero-shot conditions. This contrast suggests that while significant progress has been made in cross-modal alignment (enabling plausible answer selection from structured options), current MLLMs suffer from severe limitations in the fine-grained understanding required for images that describe structured knowledge with multi-modal entities and relational topology. Addressing this bottleneck is crucial for future advancements. Next, we will explore the details of MLLM's understanding of abstract structured knowledge even further.

%% file: chapters/5.3-figure.tex
\begin{figure*}
  \centering
  \subfigure[Results on Task 1: Entity Counting]{
  \includegraphics[width=0.9\textwidth]{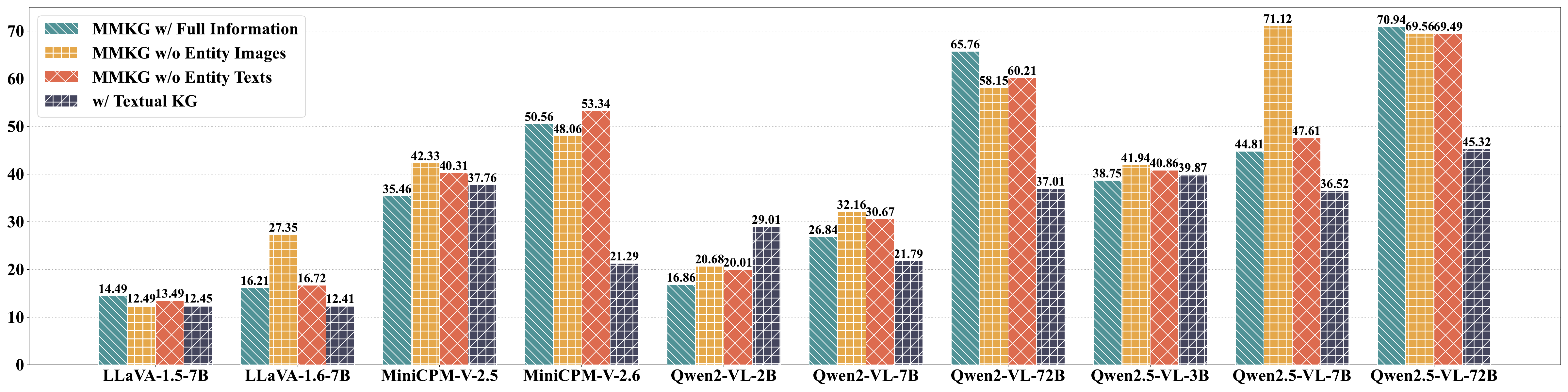}
  \label{figure::ablation1}
  }
  \subfigure[Results on Task 3: Entity Completion]{
  \includegraphics[width=0.9\textwidth]{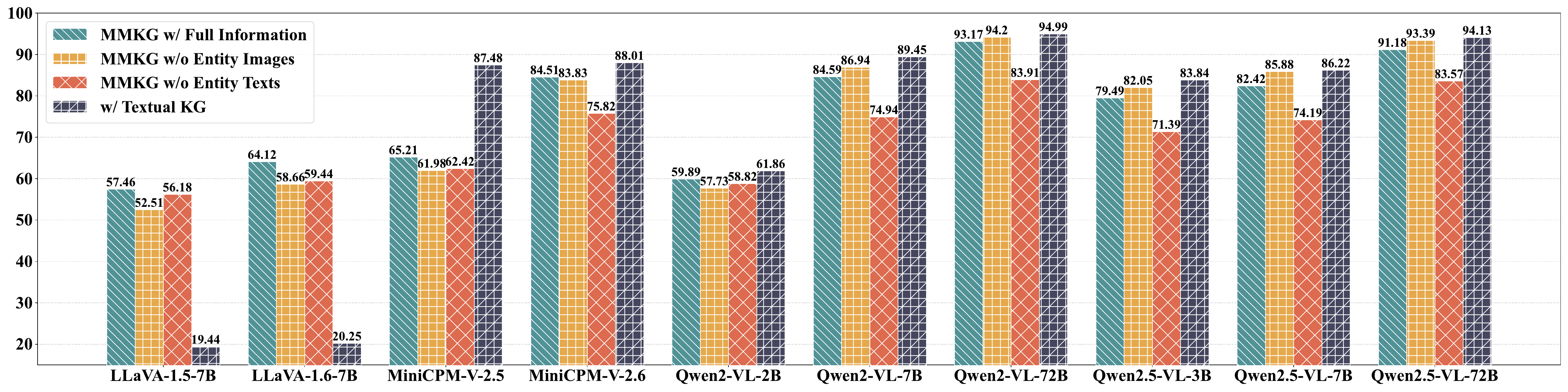}
  \label{figure::ablation2}
  }
  \vspace{-16pt}
  \caption{A further study about the modality contribution in the visual MMKG on 9 MLLMs. The visual input $\mathcal{I}$ which is a subgraph from MMKG consists of multiple modality information. We conduct this experiment to validate whether MLLMs can fully understand both the entity's images and the text descriptions.}
  \label{figure::ablation}
\end{figure*}
\begin{figure}[]
  \centering
  \includegraphics[width=\linewidth]{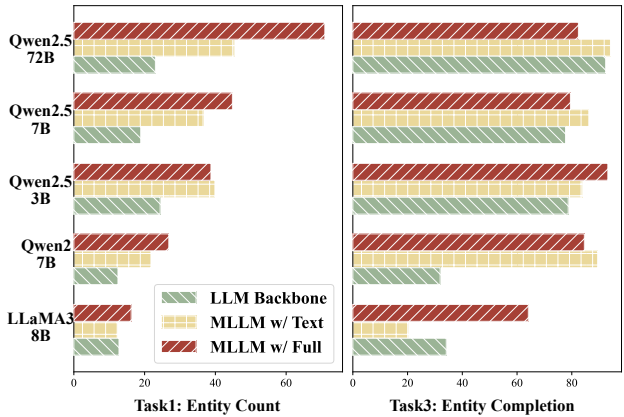}
  \vspace{-8pt}
  \caption{Results on similar backbones of the MLLMs. We aim to validate whether the extra training contributes to the abstractive understanding capability.}
  \label{figure::backbone}
  \vspace{-16pt}
\end{figure}

\input{chapters/5.4-figure}

%% file: chapters/5.4-figure.tex
\begin{figure*}[t]
  \centering
  \includegraphics[width=\linewidth]{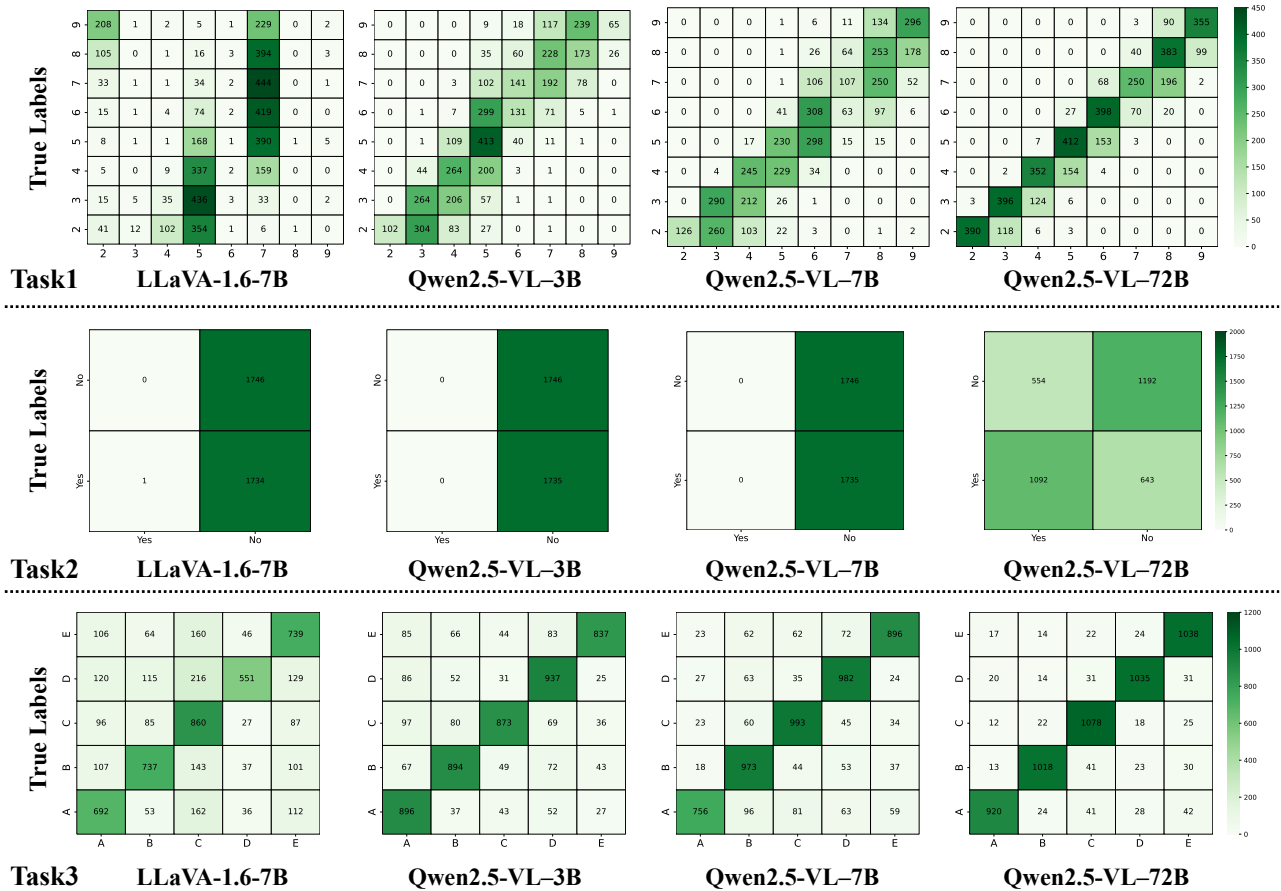}
  \vspace{-24pt}
  \caption{Confusion matrices of serveral MLLMs on the three tasks. The x-axis represents the predicted answers and y-axis represents the true label. We can observe the distrbution biases in the selected MLLMs from the matrices.}
  \label{figure::case}
  \vspace{-16pt}
\end{figure*}

%% file: chapters/5.3-ablation.tex
\subsection{Modality Contribution Analysis (RQ2)}
This section presents a deeper analysis of modality contributions within the {\benchmark} benchmark. As mentioned before, we synthesize MMKG with graph visualization APIs as the visual inputs which consist of multiple pieces of information including images and texts in the pixels, making the images semantic-rich and difficult to understand. This raises a critical research question: Do MLLMs accurately interpret and leverage all available modalities for their final predictions?
\par Therefore, we conduct modality contribution analysis. We generate the same benchmark datasets with \textbf{less informative images} as inputs, with \textbf{entity images and texts being removed}, and conduct experiments. We also conduct experiments on MLLMs with \textbf{full-text inputs} to validate whether the MMKG in visual modality is a better choice compared with text-based KG. Results in Figure \ref{figure::ablation} reveal counterintuitive phenomena. Contrary to the expectation that removing visual elements would degrade performance by limiting contextual understanding, Figure \ref{figure::ablation1} demonstrates that many MLLMs exhibit improved accuracy with simplified inputs. Notably, Qwen2.5-VL-7B achieves 159\% performance gain when entity images are removed from the whole subgraph, which outperforms 72B models. Similar patterns occur in LLaVA and MiniCPM-V architectures. Only larger models like Qwen2.5-72B align with intuitive expectations. This suggests that extraneous visual details can impede smaller MLLMs during entity enumeration tasks, where information reduction paradoxically enhances performance.

\par For completion task in Figure \ref{figure::ablation2}, such phenomena has been mitigated but still exists. More remarkably, substituting visual MMKGs with textual KG descriptions significantly boosts accuracy for MiniCPM and Qwen models. This indicates superior comprehension of textual knowledge representations compared to their visual counterparts. To some extent, this may be because FB15K-237, as well-known open-source data, may have been included in the pre-trained corpus.
\par These observations collectively indicate a fundamental limitation: Current MLLMs lack robust capabilities for abstract structural understanding of visual knowledge representations. When confronted with complex visual morphologies like multiple entities with relational topology, smaller models experience cognitive overload—manifesting as impaired quantitative reasoning. Furthermore, the consistent superiority of text-based inputs reveals that despite multimodal architectures, visual processing capabilities remain underdeveloped relative to linguistic reasoning.
\par We further examine the impact of multi-modal adaptation on underlying LLM backbones. Since MLLMs typically extend text-only LLMs through incremental multi-modal pretraining or instruction tuning, we evaluate whether these adaptations enhance MLLMs' abstractive understanding capability. We select the LLM backbones of different MLLMs (e.g. Qwen2.5-7B for Qwen2.5-7B-VL) to validate its performance on {\benchmark} with full-text prompts. As shown in Figure \ref{figure::backbone}, multi-modal adaptation generally confers significant advantages. Most MLLMs outperform their foundational LLMs even on text-only KG inputs, indicating that multi-modal training enhances both cross-modal capabilities and textual reasoning on {\benchmark}. The sole exception is Qwen2.5-72B on Task 3, where its exceptionally powerful LLM backbone potentially diminishes observable gains. Overall, these findings confirm that multi-modal extension and adaption from LLMs to MLLMs effectively augment model capabilities across abstractive understanding tasks.

%% file: chapters/5.4-case.tex
\subsection{Case Study (RQ3)}
To intuitively visualize the difference in the performance of the different models, we plotted some confusion matrices for the MLLM prediction results in Figure \ref{figure::case}. Task 1 results reveal a discernible performance stratification: stronger models like Qwen2.5-VL-72B exhibit minimal gap from ground-truth counts even when incorrect, whereas underperforming models (e.g., LLaVA) display erratic and biased answer distributions. This contrast underscores fine-grained capability gaps—competent MLLMs generate marginal errors, while weaker models demonstrate fundamental comprehension failures. 
Task 2 analyses further expose critical limitations. Most MLLMs exhibit severe prediction bias, systematically defaulting to specific output categories rather than random guessing. This suggests that these MLLMs do not make random guesses, but very biasedly fall back to one certain answer, reflecting a serious lack of anomaly detection ability of MLLMs. While Qwen2.5-VL-72B shows marginally better judgment in local anomaly detection, its overall performance remains suboptimal. Collectively, these results reveal persistent hallucination tendencies and an alarming deficiency in relational reasoning within visual KGs.

%% file: chapters/6-conclusion.tex
\section{Conclusion}
In this paper, we introduce a novel perspective for evaluating the abstractive visual understanding capabilities of MLLMs on structured knowledge with multi-modal entity information and relational topology, an area that has not been thoroughly explored before. To support this new evaluation framework, we construct the {\benchmark} benchmark with a new pipeline, which includes a diverse set of task types. Through comprehensive evaluation using the {\benchmark} benchmark, we discover significant insights from the results that suggest current MLLMs still face challenges in effectively understanding abstractive content. These findings highlight the need for further advancements in MLLM architectures and methodologies, particularly in their ability to understand and generate complex, structured relational knowledge, which is also the future direction for developing multi-modal artificial general intelligence.